\documentclass[letterpaper]{article} % DO NOT CHANGE THIS
\usepackage{aaai23}  % DO NOT CHANGE THIS
\usepackage{times}  % DO NOT CHANGE THIS
\usepackage{helvet}  % DO NOT CHANGE THIS
\usepackage{courier}  % DO NOT CHANGE THIS
\usepackage[hyphens]{url}  % DO NOT CHANGE THIS
\usepackage{graphicx} % DO NOT CHANGE THIS
\usepackage{amsmath}
\usepackage{amsfonts}
\usepackage{dsfont}
\usepackage{titlesec}
\usepackage{natbib}  % DO NOT CHANGE THIS AND DO NOT ADD ANY OPTIONS TO IT
\usepackage{caption} % DO NOT CHANGE THIS AND DO NOT ADD ANY OPTIONS TO IT
\usepackage{algorithm}
\usepackage{algorithmic}

\usepackage{newfloat}
\usepackage{listings}
\usepackage{xcolor}

\DeclareCaptionStyle{ruled}{labelfont=normalfont,labelsep=colon,strut=off} % DO NOT CHANGE THIS
\floatstyle{ruled}
\newfloat{listing}{tb}{lst}{}
\floatname{listing}{Listing}
\title{Surrogate Assisted  Monte Carlo Tree Search in Combinatorial Optimization}
\author {
    Saeid Amiri\textsuperscript{\rm 1},
    Parisa Zehtabi\textsuperscript{\rm 2},
    Danial Dervovic\textsuperscript{\rm 3},
    Michael Cashmore\textsuperscript{\rm 3}
}
\affiliations {
    \textsuperscript{\rm 1}JP Morgan AI Research. New York City, NY, USA\\
    \textsuperscript{\rm 2}JP Morgan AI Research. London, UK\\
    \textsuperscript{\rm 3}JP Morgan AI Research. Edinburgh, UK\\
    saeid.amiri@jpmchase.com, parisa.zehtabi@jpmorgan.com, danial.dervovic@jpmchase.com,  michael.cashmore@jpmorgan.com
}

\usepackage{bibentry}
\begin{document}

\maketitle

\begin{abstract}
Industries frequently adjust their facilities network by opening new branches in promising areas and closing branches in areas where they expect low profits.
In this paper, we examine a particular class of facility location problems. 
Our objective is to minimize the loss of sales resulting from the removal of several retail stores. 
However, estimating sales accurately is expensive and time-consuming.  
To overcome this challenge, we leverage Monte Carlo Tree Search (MCTS) assisted by a surrogate model that computes evaluations faster. 
Results suggest that MCTS supported by a fast surrogate function can generate solutions faster while maintaining a consistent solution compared to MCTS that does not benefit from the surrogate function.
\end{abstract}

\section{Introduction}
\label{sec:intro}
As populations shift, market trends change, and customer demands evolve, many service industries and retail stores are faced with the decision of adding, removing, relocating, or consolidating their facility locations. 
An example is a 2018 survey that showed the market trends resulting in some liquor stores becoming obsolete~\footnote{https://www.forbes.com/sites/taranurin/2019/11/22/independent-liquor-stores-will-become-obsolete-believe-nearly-one-out-of-two-owners/?sh=5eb971f57b9b}.  
In this paper, we focus on a particular class of facility location problem that involves closing a fixed number of retail stores in which computing the features of the evaluation function is expensive. 
This problem is a Combinatorial Optimization (CO).
COs are often NP-hard and computationally intractable due to the large state-spaces. 
Consequently, solving CO problems often requires designing heuristics or approximation algorithms~\cite{williamson2011design}. 
Furthermore, real-world optimization problems are often complex, nonlinear, and may have multiple objectives and constraints that can be computationally expensive to evaluate.
The solutions to CO often involves the design of heuristics or approximation algorithms.

Monte Carlo Tree Search (MCTS)~\cite{kocsis2006bandit,coulom2006efficient} is a popular technique for solving search problems in large spaces, particularly in the domain of games. 
It involves building a search tree of possible actions and their corresponding outcomes, and using evaluations (simulations) to estimate the
value of each action. 
MCTS has been applied to a wide range of problems such as games~\cite{silver2017mastering,rubin2011computer}, robotics~\cite{kim2020monte}, finance~\cite{vittori2021monte} and music~\cite{liebman2017designing}.
Recent works have used  MCTS in CO.
One work used Graph Neural Networks and Reinforcement Learning in order to compute heuristics for the MCTS-based action selection in scheduling and vehicle routing problems~\cite{oren2021solo}.
MCTS has also been applied to capacity expansion in a residency matching problem to find an optimal policy for matching medical doctors to hospital vacancies~\cite{abe2022anytime}. 

\begin{figure}[t]
  \begin{center}
    \includegraphics[width=\columnwidth]{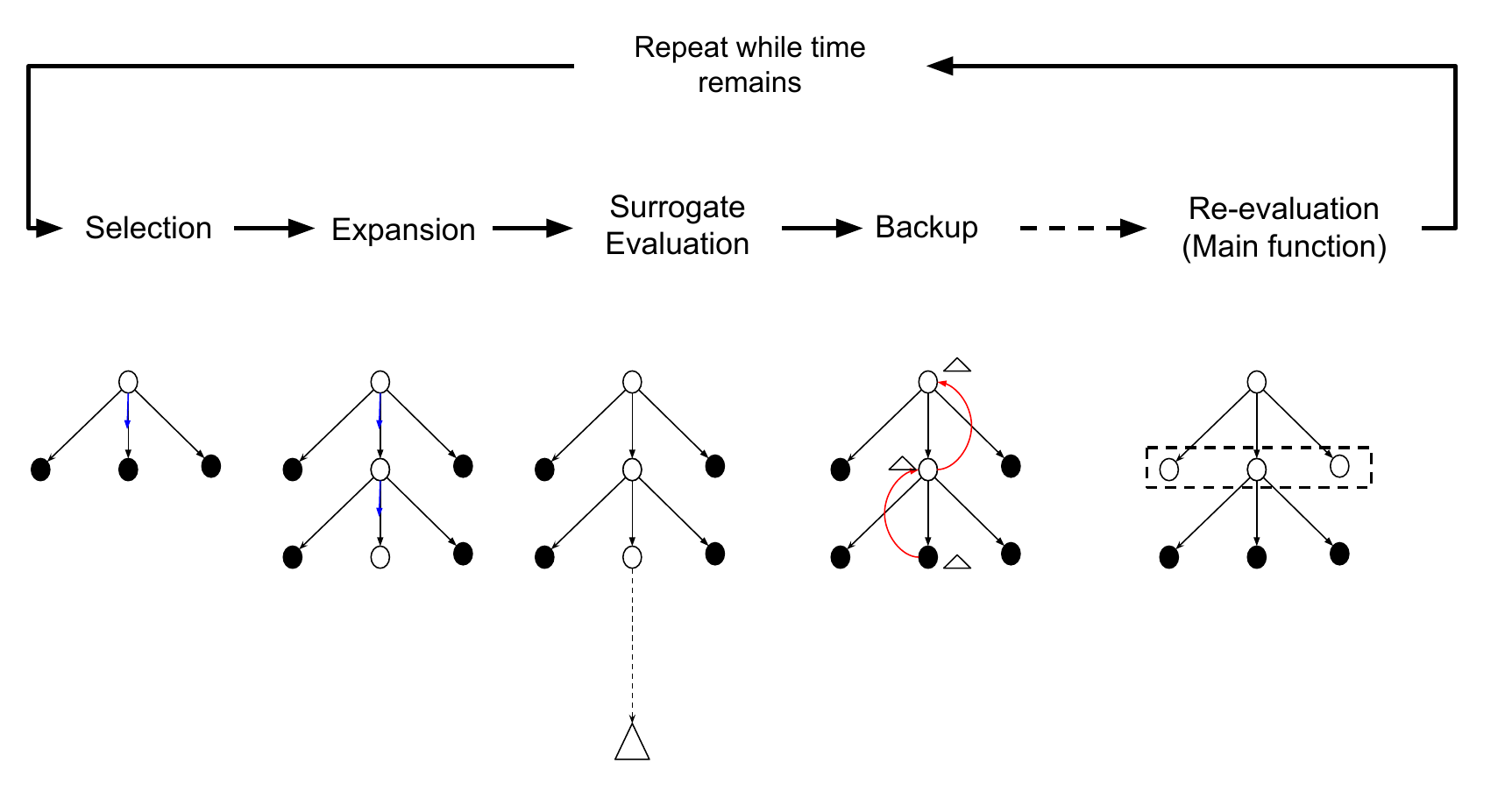}
    \caption{Surrogate assisted Monte Carlo Tree Search (SMCTS) where an occasional reevaluation step refines the node values. }
    % Figure link : https://docs.google.com/drawings/d/1sjIeCwPGB_atBQF8K994fElco2YdIDpF6ggtN2wxgdw/edit?usp=sharing
    \label{fig:mcts_general}
  \end{center}
   
\end{figure}

Inspired by these successes, we leverage MCTS in the facility location problem.
We propose Surrogate-assisted MCTS (SMCTS) to solve a combinatorial search problem where we use a fast surrogate evaluation function in concert with the slow default evaluation function. 
The main evaluation function is a regression model that evaluates the current network profitability but is computationally expensive due to the varying network-dependent features. 
The surrogate is fast to compute but is less accurate. 
The choice of an efficient surrogate function is its own research problem, and we simply assume that a surrogate function is available.
In this paper, we focus on how to use the surrogate function jointly with the main evaluation function, aiming at faster solution computation. 
Figure~\ref{fig:mcts_general} depicts the SMCTS steps where the selection, expansion, evaluation (by surrogate function), and backup is complemented by an occasional re-evaluation step that takes place in order to refine possible inaccurate surrogate model evaluations. 
%We apply our method to a liquor store network adjustment problem in which store closure should result in the minimum loss of profit.
%Empirical results on a public dataset suggest that our approach can scale for facility placement problems and can outperform greedy approaches.

We apply this approach to the problem of store closure in a network of liquor stores with the goal of minimizing the overall sales loss. 
Our empirical results show that by using MCTS with a surrogate function, the overall computation time is reduced.

\section{Related Work}
\label{sec:related}
Classes of facility location problems are among the fundamental problems in Operations Research. 
Traditionally, they have been framed using Operation Research techniques such as set covering \cite{namazian2021decision,murray2016maximal,miliotis2002hierarchical}, maximal covering ~\cite{church1974maximal,berman2002generalized}, or p-median problems where the goal is minimizing the travel distance from customers~\cite{kariv1979algorithmic}.
Facilities could be static such as branches and warehouses~\cite{zaikin2020branch} or dynamic such as charging stations~\cite{andrenacci2016demand,drezner1991facility,wesolowsky1973dynamic}.

Most works formulate the problem either as integer programming or clustering methods.
In the former, various heuristic techniques such as tabu search, Lagrangian relaxation~\cite{santos2009solving}, greedy interchange~\cite{kuehn1963heuristic}, branch and bound~\cite{jarvinen1972branch,dupont2008branch}, primal and dual~\cite{captivo1991fast}, binomic approach~\cite{maniezzo1998bionomic} and gamma heuristics~\cite{rosing1999gamma} have been used. 
In the latter, some works leveraged unsupervised clustering methods to find the demand centroids for charging stations and assign customers based on distance to the closest centroid~\cite{andrenacci2016demand,ip2010optimization}.
In some papers, Analytic Hierarchy Process have been proposed for banking branch location based on customer demographic and economic features~\cite{gorener2013application,sharmin2019bigbank}.

% https://www.retaildogma.com/why-retail-stores-are-closing/
% https://scholar.google.com/scholar?hl=en&as_sdt=0,33&as_ylo=2022&q=%22facility+location+problem%22

Most works have used the travel distance and customer demand as the only features in their objective and constraints formulation.
\citet{zaikin2020branch} set the customer dissatisfaction minimization as the goal in the branch closure problem using Max-SAT methods.

Almost all works have considered the solution to the facility location problem as a one-shot static solution where the location of multiple facilities are selected all at once. 
In practice, firms and industries would decide to alter the exisiting network of facilities. 
One work considered removing or addition of existing facilities using integer programming and approximation techniques~\cite{wang2003budget}.

In comparison, this paper focuses on a data-driven evaluation function that estimates the overall sales profit based on numerous features such as the number and the distance of nearby facilities. 
In addition, we leverage surrogate evaluations as a fast evaluation for the facility location problems.

\paragraph{Surrogate-assisted optimization}
Leveraging surrogates for optimization has already been explored by researchers in other contexts~\cite{namazi2020surrogate,gu2021surrogate,liu2013gaussian}. For instance, \cite{liu2013gaussian} proposed a Gaussian Process-assisted evolutionary algorithm to solve computationally expensive problems.
A surrogate model has been used to prune the solution search space in the Travelling Thief problem~\cite{namazi2020surrogate}.

To the best of our knowledge, this is the first work of leveraging surrogates in MCTS in CO problems.

%\paragraph{MCTS with abstractions:} Leveraging abstractions in has been %beneficial in decision making problems. 
%Abstractions can have many forms but in general it aims at coarser representation of the problem statement by designing hierarchies or leveraging macro-actions~\cite{gabor2019subgoal,amato2014planning,sutton1999between}. 
%Similarly, by using surrogate assisted MCTS, we represent the problem in a coarser resolution in order to speed up the computation.  

\section{Problem Statement}
\label{sec:problem}
This problem is a class of facility location problem in which a fixed number of retail stores are going to be closed. 
There is a city network of $N \in \mathbb{N} $ stores. 
We seek to remove $M$ stores ($ M < N$), that  result in minimum forgone sales of the network. 
Our decision variable is the vector $X~(|X|=N)$ such that:

$$X_j = \left\{
        \begin{array}{ll}
            1 & \text{if store}~j~\text{remains open} \\
            0 & \text{If store}~j~\text{remains closed}
        \end{array}
    \right.$$
and the objective is expressed as:
%$$
%\text{Minimize} \sum_{j=1}^N F_m(X_j, j) - %\sum_{j=1}^N (1-X_j)F_m(X_j, j)
%$$
$$
\text{Minimize} \sum_{j=1}^N F_m( \mathds{1}_N ,j)- \sum_{j=1}^N F_m(X,j)
$$
$$
%\text{Subject to} \sum_{j=1}^N X_j = M 
\text{Subject to}~||X||_2 = M
$$
where $\mathds{1}_N$ is a vector of $1$s and size $N$. The objective as mentioned above is to minimize the total loss of sales as a result of store closures and the constraint states that exactly $M$ stores will be closed. 
%$F_m : \{ 0, 1 \} \times \{1, \ldots,  N\} \to \mathcal{R}$, is an evaluation function %that estimates the sales of store $j$. 
$F_m$ is an evaluation function that estimates the sales of store $j$.
It is important to note that the sales estimated per store depend not only upon the features of that store, but also on other stores, including whether or not they are closed.

Next, we describe how we find the solutions to this optimization problem.

\section{Framework}
\label{sec:framework}
In this section, we explain the surrogate assisted MCTS framework.  
\paragraph{Node representation:} 
In our search tree, 
a node is identified by the set of candidate stores for removal according to the path from the root.
The root node represents no store removal and the tree depth is $M$ where $M$ is the total number of stores to be removed from the network.
In other words, the nodes at depth $M$ are terminal nodes.
Each node keeps duplicate attributes $V'_s$ and $N'_s$ for its value and the number of visits in case it goes through the re-evaluation step.
SMCTS has five components:
\begin{itemize}
    \item \textbf{Selection:} Starting at the root node, the tree policy, in our case UCB1~\cite{kocsis2006bandit},  is used to select the next node based on its value, as described below:
    $$\operatorname{argmax}\left\{v_s + C\sqrt{N_p/N_s}\right\}$$
    where $v_s$ is the value of the node $s$, $N_s$ the number of times node $s$ is visited, $N_p$ is the number of times the parent node has been visited.
    The choice of $C$ affect the ratio of exploration versus exploitation in the search.\footnote{We use $v_s'$, $N_p'$, $N_s'$  instead, after the node is reevaluated.}
    \item \textbf{Evaluate:} A node can be evaluated using functions $F_m$ and $F_s$.
    $F_m$ is the main evaluation function that is costly to compute. 
    $F_s$ is an approximate surrogate function that is faster to compute but less accurate compared to $F_m$.
    \item \textbf{Backup:} The return generated by the main or surrogate evaluation function is backed up to update the values.
    \item \textbf{Expand:} A node is expanded to its children by removing any of the remaining stores from the network. 
    The number of children expanded is equal to the remaining number of stores in the network. 
    Removing a store can be denoted as taking action $a_i \in A$ meaning the removal of the $i$th store. 
    \item \textbf{Re-evaluation:} A node's children are re-evaluated if their values are within the estimation error of the neighboring nodes in the same tier. 
\end{itemize}
Next, we briefly explain the SMCTS algorithm. 
\paragraph{Algorithm:} Algorithm~\ref{alg:MCTS} requires a surrogate function $F_s$ with the error bound $\sigma_s$, (in our case, $\sigma_s$ is the difference between the Root Mean Squared Error (RMSE) of the $F_s$ and $F_m$) and a main evaluation function is $F_m$.
%For a search problem with a costly-to-compute $F_m$, we use the surrogate function to facilitate the computational burden.    
%Algorithm~\ref{alg:MCTS} presents SMCTS.
Node $s$ is initialized with the root node $s_0$.
The selection is done using the UCB1%~\footnote{If the node has been reevaluated, UCB1 uses $V'_s$ and $N'_s$ for action selection} 
~algorithm where it suggests the best action $a$ denoting the next best store for removal (Lines~\ref{l:select}-\ref{l:select_a}).
Once the next node is selected, it expands into new children. (Lines~\ref{l:check_leaf} -\ref{l:expand}). 
%The number of child nodes of the parent node is equal to the total number of facilities in the network.
The value of the node is estimated by the surrogate function $F_s$ and backed up to the parent nodes recursively (Lines~\ref{l:evaluate}-\ref{l:backup}).
The novelty of SMCTS is in the \textbf{re-evaluate} step where an occasional refinement of node values is done in order to reduce value errors. 
The re-evaluation step is presented in Algorithm~\ref{alg:reevaluate}.
This algorithm is called when all the children of node $s$ are visited an equal number of times. 
In that case, Algorithm~\ref{alg:reevaluate} sorts the values of all children in the subtree (sharing same parent node). 
We name the values of two adjacent sorted nodes $V_{s_i}$ and $V_{s_{i+1}}$. 
These values may not be accurate as they have been evaluated using $F_s$, therefore if  $V_{s_{i+1}} - \sigma_s$ is less than   $V_{s_i} + \sigma_s$, then these node values need to be updated with $F_m$.   
The number of times that re-evaluate is called would depend on $\sigma_s$ and the distribution of node values.

\begin{algorithm}[tb]
    \caption{Surrogate-assisted MCTS}
    \label{alg:MCTS}
    \textbf{Input}: Surrogate function $F_s$, evaluation function $F_m$, action set $A$, root node $s_0$, error bound $\sigma_s$ \\
    %\textbf{Parameter}: $k$ number of iterations \\
    %\textbf{Output}: 
    
    \begin{algorithmic}[1] %[1] enables line numbers
    %\STATE $s \gets Initialize(m,n)$
        
        \WHILE{Computational budget}
        \STATE $s \gets s_0$
        \WHILE{$s.terminal$ is $False$ }
        \STATE $a \gets Select(s,A)$   \label{l:select}
        \STATE $s \gets s.children[a]$ \label{l:select_a}

        \IF{$s.leaf$ is $True$} \label{l:check_leaf}
        \STATE $Expand(s)$    \label{l:expand}
        \ENDIF
        
%        \IF{$s.terminal$ is $True$} \label{l:check-terminal}
        \STATE $ v \gets Evaluate(s,F_s)$ \label{l:evaluate}
        \STATE $Backup(s,v)$ \label{l:backup}
        %\STATE Break            
%        \ENDIF
        \IF{$s.leaf$ is $False$ and $s.children$ equally visited}
        \STATE $Re-evaluate(s,F_m, \sigma_s)$
        \ENDIF
        %\STATE \textbf{Break}
        %0\ELSE
%        \STATE \saeidrevise{should the two lines below go to the top of the loop?}

                \ENDWHILE
        \ENDWHILE
        %\STATE \textbf{return} Top $m$ nodes
\RETURN Node with the highest value
    \end{algorithmic}
\end{algorithm}

\begin{algorithm}[tb]
    \caption{Re-evaluate nodes }
    \label{alg:reevaluate}
    \textbf{Input}: Node $s$, $F_m$, $\sigma_s$ error bound \\
    
    \begin{algorithmic}[1] %[1] enables line numbers
    %\STATE $s \gets Initialize(m,n)$
        \STATE Sort children of $s$ based on value
%        \IF{children of $s$ have been visited equally}
        \FOR{$i$ in $[0, s.children.length-1]$}
        \STATE $v_{s_i} \gets s.children[i].value $
        \STATE $v_{s_{i+1}} \gets s.children[i+1].value $
       % \STATE $\Delta \gets V_{s_{i+1}} - V_{s_i} $
        \IF{$v_{s_{i+1}} - \sigma_s  < v_{s_i} +\sigma_s$}
        \STATE  $ v'_{s_{i+1}} \gets Evaluate(s.children[i+1],F_m)$
        \STATE  $ v'_{s_{i}} \gets Evaluate(s.children[i],F_m)$
        \STATE $Backup(s_i,v'_{s_i})$
        \STATE $Backup(s_{i+1},v'_{s_{i+1}})$
        \ENDIF
        \ENDFOR
        
%        \ENDIF
    \end{algorithmic}
\end{algorithm}

%
%\begin{figure}[t]
%  \begin{center}
%%    \includegraphics[width=0.7\columnwidth]{Figures/marshalltown.PNG}
 %   \caption{The liquor store locations in Marshalltown, IA 
 %   }
 %   \label{fig: sample_town}
 % \end{center}
   
%\end{figure}

\section{Experiments}
\label{sec:exp}
This section details the conducted experimental evaluation, analysing the performance of SMCTS in different problem settings.
Our evaluation aimed at studying the following hypothesis:
\begin{enumerate}
    \item In scenarios with higher scale (large number of total stores or large number of removals), SMCTS tends to leverage surrogate function more than the evaluation function.
    \item The number of surrogate evaluations depends on the surrogate quality. 
    The higher error it has, the more re-evaluation steps are needed.
    \item With an efficient choice of a surrogate function, SMCTS maintains  a solution consistent with unassisted MCTS.
\end{enumerate}

Next we explain the dataset and the evaluation functions.
%It is important to note that this deduction is dependent on the choice of the surrogate function.

\paragraph{Dataset:} We use the Iowa Liquor Dataset~\footnote{https://data.iowa.gov/Sales-Distribution/Iowa-Liquor-Sales/m3tr-qhgy} that contains the daily purchase information of various liquors in each store in the state of Iowa. 
The dataset has the information of the stores such as the store name, address, coordinates, zip code with $978$ unique values, and the city name with $476$ cities and the type and amount of the liquor they have sold at each day.
\begin{figure}[t]
  \begin{center}
    \includegraphics[width=0.8\columnwidth]{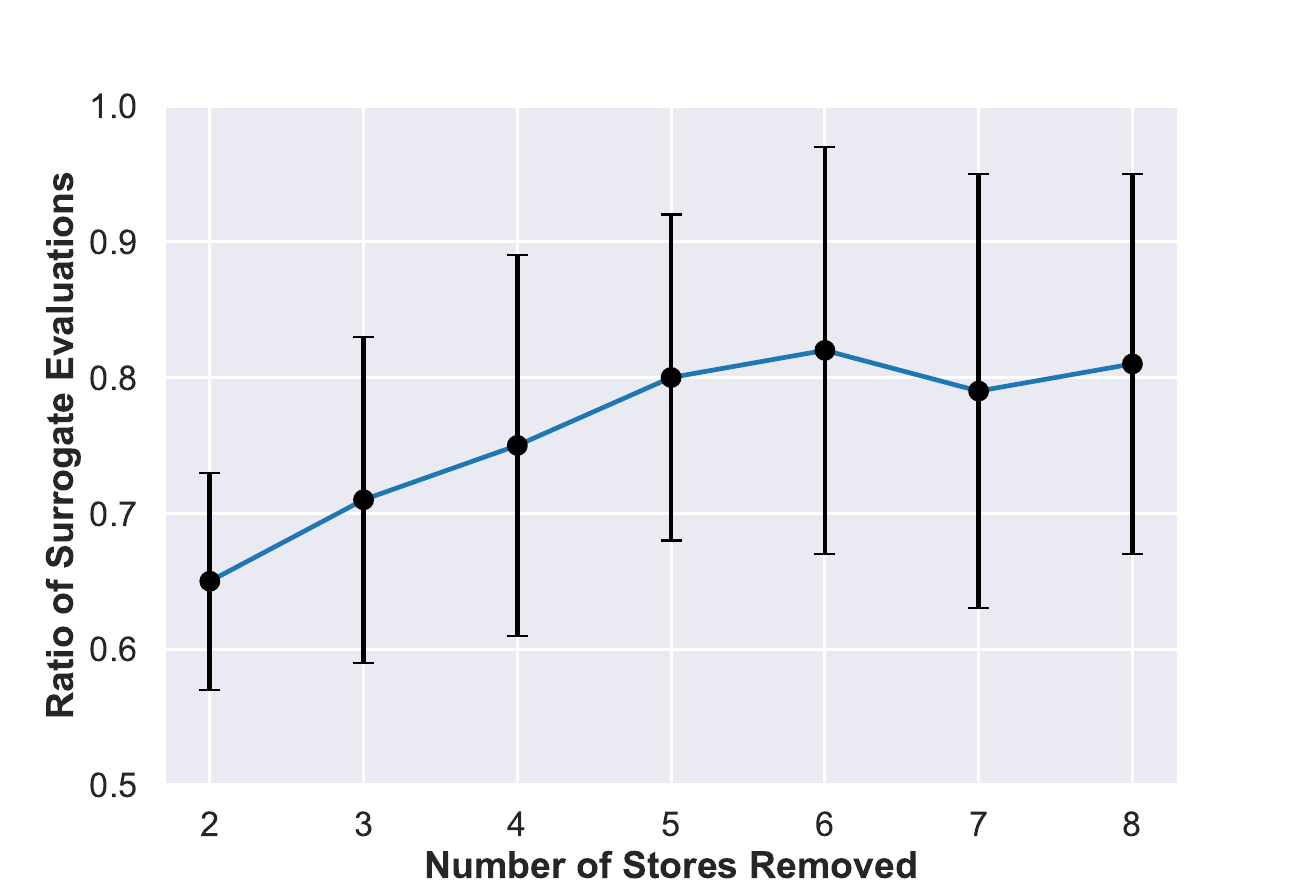}
    \caption{SMCTS where an occasional reevaluation step refines the node values. The horizontal axis represents the number of stores that need to be removed. %With the increase in the removed stores, the ratio of using $F_s$ increases, resulting in faster computation.
    }
    % Figure link : https://docs.google.com/drawings/d/1sjIeCwPGB_atBQF8K994fElco2YdIDpF6ggtN2wxgdw/edit?usp=sharing
    \label{fig:exp_surrogate_ratio}
  \end{center}
   
\end{figure}
\begin{figure}[t]
  \begin{center}
  \vspace{-0.5mm}
    \includegraphics[width=0.8\columnwidth]{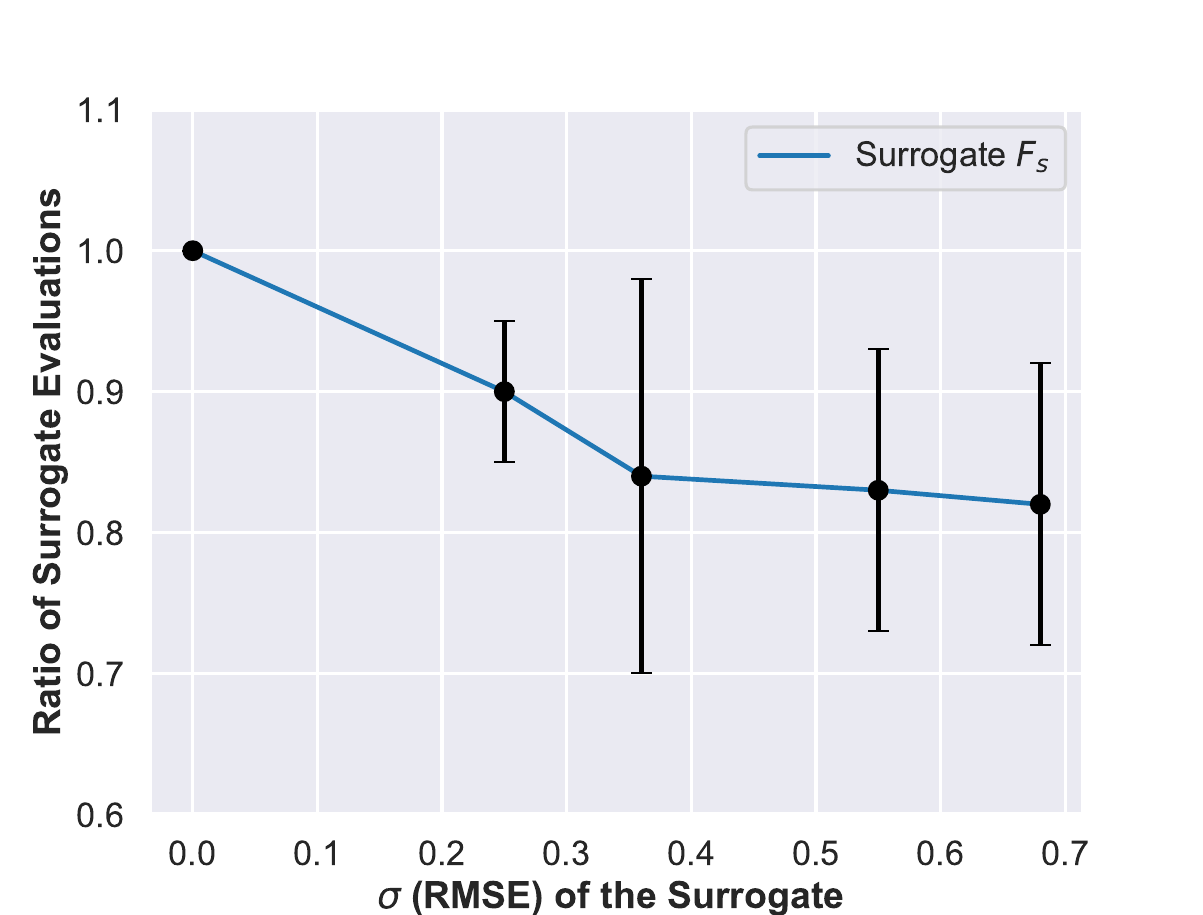}
    \caption{SMCTS with various surrogate errors. The vertical axis is the ratio of surrogate function $F_s$ evaluations to the total evaluations. 
    The horizontal axis represent surrogate functions with increasing normalized RMSEs.}

    \label{fig:error}
  \end{center}
   
\end{figure}
\begin{figure}[t]
  \begin{center}
    \includegraphics[width=\columnwidth]{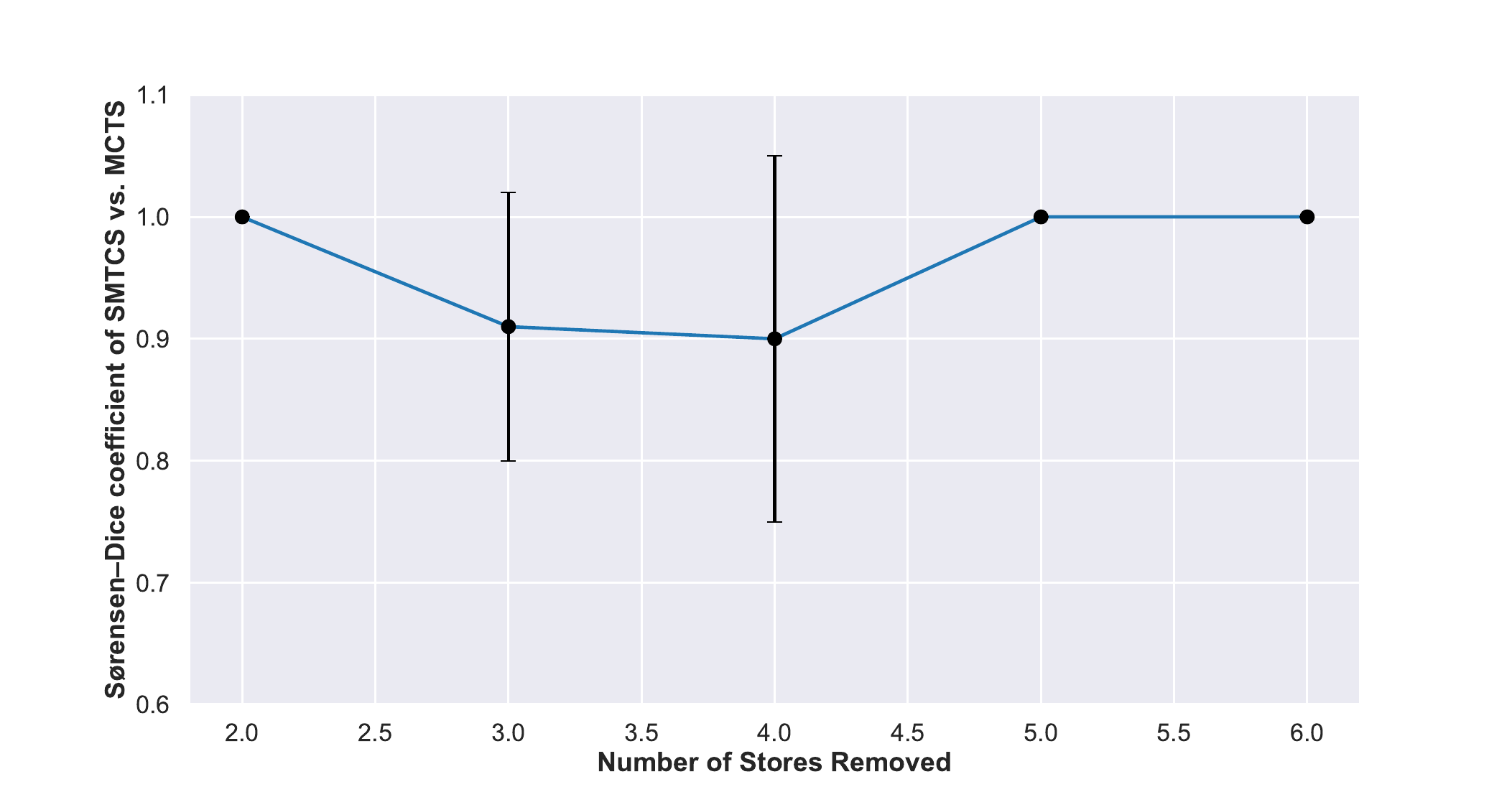}
    \caption{Evaluation of the consistency of stores selected by SMCTS vs. MCTS. The vertical axis shows the number of the selected stores by SMCTS being different from MCTS. The results are the average of $10$ counties that are randomly selected. }

    \label{fig:consistency}
  \end{center}
   
\end{figure}
We preprocess the data by calculating the total sales of all liquors at each store in a year and by defining some new features such as the number of stores in $0.5$ miles proximity.
These additional features are network dependent and require to be recomputed for every node in the tree, resulting in $F_m$ to be costly.

\paragraph{Evaluation Functions:}
Our main evaluation function $F_m$ is an XGBoost regression model that estimates the sales amount for a store. 
Given a store removal, some features in the dataset need to be recalculated, resulting in a costly evaluation. 
To create a surrogate function $F_s$, we use a subset of the features of the dataset and train another XGBoost regression model. 
$F_s$ is less accurate on sales estimation compared to $F_m$. 
In our case, $F_s$ has a normalized RMSE of $0.27$ and $F_m$ has the RMSE of $0.16$ (both on the test set). %but $F_m$ is on average $X$ times faster in computation of the sales. 
We use SMCTS for liquor store removals for a given county with varying counts of store to remove. 
Figure~\ref{fig:exp_surrogate_ratio} shows the average ratio of the number of times $F_s$ has been called versus $F_m$ for ten counties (randomly sampled) where the number of stores in those counties ranged from $17$ to $64$, representing $70$ store removal problems.  
We observe that with the increase in the number of removed branches, the relative number of times the surrogate function is called increases, facilitating reduction of the overall evaluation burden.  
Figure~\ref{fig:error} shows the ratio of surrogate evaluation to the total evaluation using various surrogates with various error bound. 
With the increase in the error of the surrogate, we observe an increase in the re-evaluation step. 
Such increase is valuable as long as SMCTS is consistent with MCTS store selection. 
Figure~\ref{fig:consistency} presents the consistency comparison of the two approaches for various store removals. 
We use the Sørensen–Dice coefficient to measure the similarity of the results of the two methods.
The values are the average of ten counties, randomly sampled from the dataset.
We observe that in most cases, SMCTS output is consistent with MCTS. 
There is a bit of inconsistency for 3 and 4 branch removals, such inconsistencies are due to the weaker estimations of $F_S$ in outlier counties.

\section{Conclusion \& Future Work}
In this work, we proposed MCTS search with surrogate functions for combinatorial optimization. 
We demonstrated that by using less accurate but faster surrogate function, we can solve optimization problems more efficiently. 
We applied our approach to a store closure problem in which the goal is to minimize the total sales loss of a retail store.

In this paper, we assumed the surrogate function is provided while this is not the case in practice. 
For future work, we propose to investigate ways to implement and design the surrogate function and the criteria for it to improve the SMCTS.
In addition, we will explore the applicability of SMCTS with other datasets and domains with stochasticity in the action space.

{\footnotesize
\paragraph{Disclaimer.}
This paper was prepared for informational purposes by
the Artificial Intelligence Research group of JPMorgan Chase \& Co. and its affiliates (``JP Morgan''),
and is not a product of the Research Department of JP Morgan.
JP Morgan makes no representation and warranty whatsoever and disclaims all liability,
for the completeness, accuracy or reliability of the information contained herein.
This document is not intended as investment research or investment advice, or a recommendation,
offer or solicitation for the purchase or sale of any security, financial instrument, financial product or service,
or to be used in any way for evaluating the merits of participating in any transaction,
and shall not constitute a solicitation under any jurisdiction or to any person,
if such solicitation under such jurisdiction or to such person would be unlawful.
}

{\footnotesize
\bibliography{aaai23}

}

\end{document}